\documentclass[letterpaper]{article}
\usepackage{aaai}
\usepackage{times}
\usepackage{helvet}
\usepackage{courier}
\usepackage{amsmath, amstext, amsfonts}
\newcommand{\mfoilp}{\textsf{mfoilp}}
\begin{document}
\title{First-order integer programming for MAP problems}
\author{James Cussens\\
Department of Computer Science\\
University of York,
York, YO10 5GE, UK\\
}
\maketitle

\section{Logic programs for MIP generation}
\label{sec:using}

Finding the most probable (MAP) model in SRL frameworks such as Markov
logic \cite{richardson05:_markov_logic_networ} and Problog
\cite{DBLP:journals/corr/abs-1304-6810} can, in principle, be solved
by encoding the problem as a `grounded-out' mixed integer program
(MIP). However, useful first-order structure disappears in this
process motivating the development of first-order MIP approaches. Here
we present \mfoilp{}, one such approach. Since the syntax and
semantics of \mfoilp{} is essentially the same as existing approaches
\cite{gordonHD09,kerstingMT14} we focus here mainly on implementation
and algorithmic issues. We start with the (conceptually) simple
problem of using a logic program to generate a MIP instance before considering
more ambitious exploitation of first-order representations.

A MIP instance consists of variable and
linear constraint declarations. It is straightforward to effect these
declarations as a logic program. In the case of \mfoilp{} this logic
program is written in the Mercury language \cite{zoltan96:_mercur} as follows.

Mercury programs are made up of modules. Mercury modules are composed
of an interface and an implementation, where everything in a modules's
implementation is hidden from other modules.  Let the Mercury module
defining a MIP instance be called \texttt{prob.m}. Its interface is as
follows:
\begin{verbatim}
:- import_module mfoilp.

:- type atom.

:- pred variable(atom::out) is multi.
:- pred constraint(lincons::out) is nondet.
:- func objective(atom) = float.
:- func lb(atom) = float.
:- func ub(atom) = float.
:- func vartype(atom) = vartype.
\end{verbatim}
Terms of type \texttt{atom} correspond to MIP variables. This type is
an \emph{abstract type}; its definition, which the user has to provide
in the module implementation, is hidden from other modules. The 4
functions \texttt{objective}, \texttt{lb}, \texttt{ub} and
\texttt{vartype} have to provide, respectively, the objective
coefficient, the lower bound, upper bound and MIP variable type
(integer- or real-valued) for
each \texttt{atom}. 
% The type \texttt{vartype} is defined in (the
% interface of) the imported module \texttt{mfoilp} as follows:
% \begin{verbatim}
% :- type vartype ---> binary ; integer 
%      ; implint ; continuous.
% \end{verbatim}
The predicate \texttt{variable/1} is used to generate all MIP
variables by Mercury's equivalent of \texttt{findall}. For example,
suppose the type \texttt{atom} were defined as follows
\begin{verbatim}
:- type protein ---> p1;p2.
:- type location_id ---> l1;l2.
:- type atom --->
  location(protein,location_id)
  ; interaction(protein,protein).
\end{verbatim}
then we could have:
\begin{verbatim}
variable(location(Protein,Location_id)) :- 
  protein(Protein), location_id(Location_id).
variable(interaction(Protein1,Protein2)) :- 
  protein(Protein1), protein(Protein2).
\end{verbatim}
where \texttt{protein/1} and \texttt{location\_id/1} generate proteins
and location\_ids, respectively.

Constraints are similarly straightforward. The \texttt{lincons} (linear
constraint) type
is defined in the \texttt{mfoilp} module as follows:
\begin{verbatim}
:- import_module prob.
:- type lterm ---> (float * atom).
:- type lexp == list(lterm).
:- type lb ---> finite(float) ; neginf.
:- type ub ---> finite(float) ; posinf.
:- type lincons ---> lincons(lb,lexp,ub).
\end{verbatim}
So a linear term is just an \texttt{atom} together with its
coefficient, and a linear constraint is just a list of such terms
(representing their sum) bounded by a lower and upper bound. Absent
bounds are represented by constants representing either positive or
negative infinity. Note that the \texttt{prob} module must be imported
so that the (abstract) type \texttt{atom} is available.
The predicate \texttt{constraint/1} should be defined so that all
required constraints can be generated by
\texttt{findall}. Fig~\ref{fig:cons} provides an example.
\begin{figure}
  \centering
\begin{verbatim}
constraint(lincons(finite(1.0),
    [1.0 * location(P1,L1), 
     1.0 * interaction(P1,P2)],posinf))  :-
  protein(P1), protein(P2), not P1 = P2, 
  location_id(L1).
\end{verbatim}
  \caption{A first-order linear constraint}
  \label{fig:cons}
\end{figure}

\section{\mfoilp}
\label{sec:mfoilp}

\mfoilp{} consists of a C program (\texttt{main.c}), a Mercury program
(\texttt{mfoilp.m}) and a Makefile. C code in \texttt{main.c} asks for
(and receives) the list of variables and list of constraints from
\texttt{mfoilp.m} which in turn gets them from \texttt{prob.m} which
defines the problem instance. It then calls the SCIP
\cite{Achterberg2009} system to create and solve the MIP instance. The
components of \mfoilp{} are connected as follows.
\begin{verbatim}
SCIP - main.c - mfoilp.m - prob.m
\end{verbatim}
Solving is invoked with
\texttt{make solution}. This causes \texttt{prob.m} to be compiled
(and linked ) to generate a problem-specific executable. This
executable is then run to solve the MIP.

\section{Branch-price-and-cut}
\label{sec:scaling}

A logic program is obviously a very flexible way of defining a MIP
instance.  However, the approach just outlined will fail with large
numbers of variables and/or constraints since all have to be squeezed
into memory prior to any solving. An alternative is to implement a
\emph{branch-price-and-cut (BPC)} approach to solving the MIP. Assume
wlog that we are minimising. In BPC some initial variables and
constraints are created as normal and the LP solution $x^*$ computed,
giving a lower bound on an optimal solution. Next a \emph{cutting
  plane} algorithm looks for valid inequalities which $x^*$
violates. If any are found they are added and a new LP solution with a
tighter (higher) lower bound can be computed. In addition a
\emph{pricer} algorithm is run to look for variables which, if
created, would produce a looser (lower) lower bound. Such variables
are those with negative \emph{reduced cost}, a quantity which can be
computed from the variable's objective coefficient and the solution to
the dual LP. If the pricer can establish that there are no variables
with reduced cost then we have a global lower bound without creating
all possible variables. If we reach such a point and all integer
variables happen to have integer values in the LP solution then the
MIP is solved. Otherwise BPC \emph{branches} on a variable to create
sub-problems and applies BPC recursively until an guaranteed optimal
solution is found.

It is easy to augment \mfoilp{} with a predicate implementing a
cutting plane algorithm:
\begin{verbatim}
:- pred cut(lpsol::in,
            lincons::out) is nondet.

cut(LPSol,CP) :-
  constraint(CP),
  activity(LPSol,CP,Activity),
  violates_bounds(Activity,CP).
\end{verbatim}
Mercury's default execution algorithm processes goals left to right,
so in this approach (1) a potential cutting plane is generated, (2)
the value of its linear expression for the given LP solution (its
\emph{activity}) is computed and (3) the predicate succeeds if this
value exceeds one of the constraint's bounds.

Similarly, a pricer can be implemented as follows:
\begin{verbatim}
:- pred price(duallpsol::in,
              variable::out) is nondet.

price(DualLPSol,Var) :-
  variable(Var),
  reduced_cost(DualLPSol,Var,RedCost),
  RedCost < 0.
\end{verbatim}

Such an approach avoids the need to generate all variables and
constraints ahead-of-time, since only those which affect the LP
bound are generated. And we can leave SCIP to take care of branching.
However, both \texttt{cut/2} and \texttt{price/2} are hopelessly
inefficient---being pure generate-and-test.

An attractive alternative is to (1) take advantage of any special
structure in variable/constraint definitions and (2) effect a
source-to-source transformation similar to \emph{unfolding} to produce
more efficient code. For example, all the variables in the constraint
in Fig~\ref{fig:cons} have positive coefficients, which can be
exploited to \emph{automatically} produce the following specialised
\texttt{cut/2} clause:
\begin{verbatim}
cut(LPSol,CP) :-
 protein(P1), location_id(L1),
 get_val(LPSol,location(P1,L1),Val1),
 Val1 < 1,
 interaction(P1,P2), not P1 = P2,
 get_val(LPSol,interaction(P1,P2),Val2),
 Val1 + Val2 < 1.
 CP = lincons( ... ). %lit abbreviated
\end{verbatim}

\mfoilp{} is currently being extended in this direction with a SCIP
\emph{constraint handler} and SCIP \emph{pricer} which will call
Mercury code to generate the required constraints and variables. (An
earlier version of \mfoilp{} called \texttt{foz} does implement
cutting planes but does not use Mercury and has a naive cutting plane
algorithm.)

\section{Related work}
\label{sec:related}

In \mfoilp{} there is a bijection between MIP variables and
first-order \emph{terms}. However, these terms can be usefully thought
of as atomic formula (and so the predicates become meta-predicates) in
which case the syntax and semantics of \mfoilp{} are basically the
same as first-order programming (FOP) introduced by \cite{gordonHD09}
where ``A MILP variable corresponds to a FOP ground atom''.  Further
work is needed to see how the approach to inference in FOP
presented in \cite{zawadzkiGP11} relates to the BPC method advocated here.

If all variables are continuous then an \mfoilp{} instance is
equivalent to a relational linear program (RLP) \cite{kerstingMT14}
which is a ``declarative LP template defining the objective and the
constraints through the logical concepts of objects, relations and
quantified variables.''  It follows that, even with integer variables,
\mfoilp{} should take advantage of the `lifted' LP solving technique
of Kersting \emph{et al} since solving linear relaxations plays such a
key role in MIP solving.

\section*{Acknowledgements}

This work was supported by a Senior Postdoctoral Fellowship SF/14/008
from KU Leuven and by UK NC3RS grant NC/K001264/1.

\bibliographystyle{aaai}
\bibliography{jc,biblio}

\end{document}